\documentclass{article}
\usepackage{spconf,amsmath,graphicx,hyperref,booktabs}
\usepackage{listings}
\usepackage{algorithm}
\usepackage{algpseudocode}
\usepackage{amsfonts}

\usepackage[table]{xcolor}
\usepackage{verbatim}
\definecolor{highlight}{HTML}{e1ffbb}
\definecolor{codegreen}{rgb}{0,0.6,0}
\definecolor{codegray}{rgb}{0.5,0.5,0.5}
\definecolor{codepurple}{rgb}{0.58,0,0.82}
\definecolor{backcolour}{rgb}{0.95,0.95,0.92}

\makeatletter
\let\oldthebibliography\thebibliography
\def\thebibliography#1{%
  \oldthebibliography{#1}%
  \fontsize{9pt}{10pt}\selectfont    
  \setlength{\itemsep}{0pt plus 0.2pt}%
  \setlength{\parskip}{0pt}%
  \setlength{\parsep}{0pt}%
}
\makeatother

\title{A MULTIMODAL DEPTH-AWARE METHOD FOR EMBODIED REFERENCE UNDERSTANDING}
\name{\parbox{\linewidth}{\centering Fevziye Irem Eyiokur\textsuperscript{1,2} \qquad Dogucan Yaman\textsuperscript{1,2} \qquad Hazım Kemal Ekenel\textsuperscript{3} \qquad Alexander Waibel\textsuperscript{1,2,4} }}

\address{\textsuperscript{1} Karlsruhe Institute of Technology, \textsuperscript{2} KIT Campus Transfer GmbH (KCT), \\ \textsuperscript{3} Istanbul Technical University, \textsuperscript{4} Carnegie Mellon University  }

\begin{document}
\maketitle
\begin{abstract}
Embodied Reference Understanding requires identifying a target object in a visual scene based on both language instructions and pointing cues. 
While prior works have shown progress in open-vocabulary object detection, they often fail in ambiguous scenarios where multiple candidate objects exist in the scene.
To address these challenges, we propose a novel ERU framework that jointly leverages LLM-based data augmentation, depth-map modality, and a depth-aware decision module. 
This design enables robust integration of linguistic and embodied cues, improving disambiguation in complex or cluttered environments. 
Experimental results on two datasets demonstrate that our approach significantly outperforms existing baselines, achieving more accurate and reliable referent detection.
\end{abstract}
\begin{keywords}
Embodied reference understanding, pointing target detection, multimodal learning
\end{keywords}
\section{Introduction}
\label{sec:intro}

Employing gaze and pointing along with verbal instruction for focus of attention has been proposed in prior work~\cite{stiefelhagen1999gaze, stiefelhagen2001estimating, stiefelhagen2004natural, waibe112005chil}.
Later, in \cite{chen2021yourefit}, a benchmark dataset along with a transformer-based approach is introduced.
The task is formulated as Embodied Reference Understanding (ERU)~\cite{chen2021yourefit}, which is identifying a specific object in a visual scene based on language instructions and pointing cues within the image. 
This task plays a key role in real-world applications such as human-robot interaction and assistive robotics where systems must determine which object a person is referring to.
While open vocabulary large models~\cite{liu2024grounding, steiner2024paligemma, bai2025qwen2} have made significant progress in detecting objects mentioned in natural language, they often fall short on the ERU, particularly in ambiguous scenes. 
When multiple instances of the same object type are present, these models tend to detect all matching candidates without the ability to disambiguate and correctly identify the pointed one.  
Moreover, when the textual instruction itself is vague or ambiguous, these models struggle even further, often failing to identify the correct object since they generally cannot utilize pointing cue.
Specifically, Large Multimodal Models (LMMs) suffer from outputting bounding boxes (bboxes) coordinates with confidence scores or sometimes even outputting incorrect format.
Besides, finetuning them is prohibitively expensive, and their inference is computationally costly. 
These limitations highlight the need for additional disambiguation cues, specifically embodied gesture signals, that can resolve referential ambiguity and enable accurate identification of the referent. 
Therefore, both embodied gesture signals and language references are crucial for identifying the referent. 
In ERU, two main challenges arise: (1) identifying candidate objects from text, as in standard grounding tasks, and (2) interpreting pointing cues from visual input to guide the final prediction. 
The latter requires understanding human pose, inferring pointing direction, and handling visual complexities such as perspective, occlusion, and depth.

\begin{figure*}[ht!]
    \centering
    \includegraphics[width=0.95\linewidth]{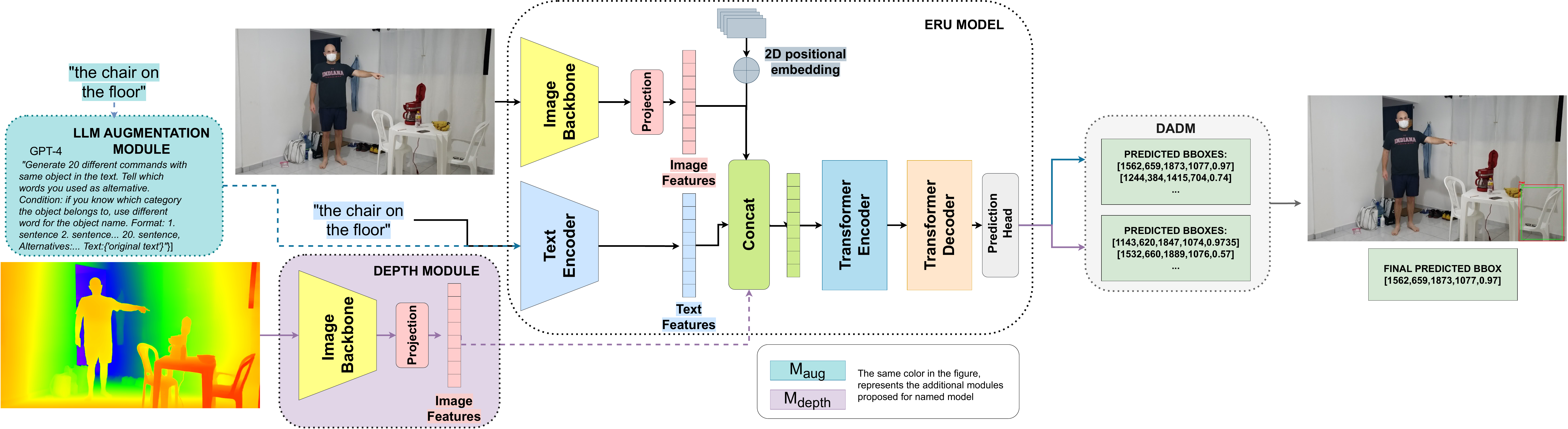}
    \caption{Overall framework. Depth and LLM Augmentation modules are used interchangeably with proposed parallel models.}
    \label{fig:model}
\end{figure*}

Earlier works utilize gaze and pointing cue to identify focus of attention \cite{stiefelhagen1999gaze, stiefelhagen2001estimating, stiefelhagen2004natural, waibe112005chil}. 
Later, the ERU task, which incorporates nonverbal cues to resolve referential ambiguity, was formulated with proposed YouRefIt benchmark in \cite{chen2021yourefit}, where saliency and pose features were used to capture pointing direction, and was later enhanced with depth-based reasoning \cite{shi2022spatial} and “virtual-touch-lines” connecting eye and fingertip \cite{li2023understanding}. 
While \cite{oyama2023exophora} uses the pointing cue to improve the detection performance of referent, \cite{constantin2022interactive, constantin2023multimodal} employ pointing detection in a cascaded system for multimodal robot dialog to detect referent.
More recent works adopt transformer-based multimodal detectors extend ERU into 3D embodied settings through benchmarks such as ScanERU \cite{lu2024scaneru} and Ges3ViG \cite{mane2025ges3vig}.   
Moreover, Referring Expression Comprehension (REC)  \cite{kazemzadeh2014referitgame,yu2018mattnet,gao2024clip,liu2024grounding,gu2021open,zareian2021open,beyer2024paligemma} can also be relevant. 
Despite these advances, most ERU methods still overlook the explicit modeling of the pointing line and depth, leaving text as the dominant cue. 
Consequently, they struggle in ambiguous or cluttered scenes where pointing and depth information could provide disambiguation. 
In this paper, we propose a novel approach to improve ERU by leveraging complementary training strategies and a depth-aware ensemble mechanism. 
First, we apply text data augmentation during training to enhance the model’s generalization capacity. 
Second, we incorporate estimated depth maps as additional input, as depth information can provide useful spatial cues. 
However, we observe that combining depth with augmentation in a single model does not yield consistent improvements, since depth can both help and hinder performance depending on the scene. 
To address this, we train two parallel models: one with augmented text data and one with normal text (no augmentation) but with depth input. 
Finally, we introduce a depth-aware decision module (DADM) that integrates the predictions of the two models. 
Specifically, the final output is selected as the bounding box closest to the predicted pointing line out of the predicted bounding boxes from two models, enabling more robust and accurate target object detection.
Our contributions are as follows.
(1) We propose LLM-based text augmentation for ERU, significantly improving target detection.
(2) We introduce a depth-aware ERU model to address failures caused by missing depth cues.
(3) We design a depth-aware decision module that combines augmented and depth-aware models for robust predictions.
(4) We achieve state-of-the-art results on two benchmarks, supported by extensive ablation studies.

\section{Methodology}
\label{sec:method}

\noindent\textbf{Text Data Augmentation.}
For each target object, we prompt GPT-4~\cite{achiam2023gpt} to generate 20 alternative sentences by replacing the object with semantically similar words only in the training set.
Although the images remain unchanged, pairing each image with 20 additional sentences increases the training set from its original size to 21 times larger. 
This strategy improves model robustness to variation in complex text prompts. 

\noindent\textbf{Depth Map Estimation.}
We employ the Depth Pro~\cite{bochkovskii2024depth} for depth map estimation. 
It's built on a multi-scale vision transformer, and effectively captures global context for accurate relative depth while preserving fine details for sharp boundaries. 
We select Depth Pro for its ability to generate seamless depth predictions directly from RGB input.

\subsection{Model Architecture}
We propose two complementary models for performing ERU. 
The first model, $M_{aug}$, is trained on augmented data, whereas the second model, $M_{depth}$, is trained on the non-augmented data with the estimated depth map modality.
Fig. \ref{fig:model} shows both of our models together with the final decision module. 

\noindent\textbf{Problem Definition.} Given an RGB image $x_{img} \in \mathbb{R}^{3 \times H \times W}$ and a text input $x_{text} \in \mathbb{N}^L$, the goal is to predict a bounding box $x_{bbox} \in \mathbb{R}^4$ that indicates an object referenced by the text instruction and pointing gesture.

\noindent\textbf{Encoders.} In both models, we use a pretrained ROBERTa encoder, a robustly optimized variant of BERT, to process the textual instruction and generate text embeddings.
For the image encoder, we employ a ResNet-101~\cite{he2016deep} to extract image features, $F_I \in \mathbb{R}^{2048 \times 8 \times 8}$.
In $M_{depth}$, a lightweight ResNet-18~\cite{he2016deep} encodes the depth map into features $F_{depth} \in \mathbb{R}^{256 \times 8 \times 8}$, as depth contains less information than the image and requires a smaller representation.
This design choice also improves computational efficiency.

\noindent\textbf{Transformer Encoder-decoder.}
We obtain features from text, image, and depth encoders, where depth applies only to $M_{depth}$.
Each embedding is projected to $256$ channels using a $1 \times 1$ convolution, after which the spatial dimensions are flattened into token sequences. 
These sequences are concatenated to form the multimodal input, which is processed by a transformer encoder. 
The encoder output, together with a set of learnable queries, is passed to a transformer decoder that produces the final object and gesture embeddings.

\noindent\textbf{Prediction Head.}
The object and gestural embeddings generated by the transformer decoder are passed to prediction heads implemented as MLPs. 
These heads output candidate bboxes, their center points for the referent, and eye-fingertip keypoints. 
For each model, predictions are ranked by confidence score in descending order, and the top two are forwarded to DADM.

\noindent\textbf{Training Details.}
Overall objective function is:
\begin{equation}
    L = \lambda_1 L_{bb} + \lambda_2 L_a + \lambda_3 L_g + \lambda_4 L_{t} + \lambda_5 L_{c}
\end{equation}
Here, $L_{bb}$ is the bbox loss (L1 + GIoU), $L_g$ represents the gestural loss, the distance between predicted and GT eye and fingertip coordinates. $L_t$ and $L_c$ are soft token and contrastive losses, respectively, following \cite{kamath2021mdetr}.
The alignment loss, $L_a$, penalizes misalignment between the eye–fingertip and the eye–object vectors by comparing their cosine similarity to the ground truth, encouraging predictions aligned with pointing.
We found best coefficients respectively $= (2,1,10,1,1)$ .

\noindent\textbf{Pointing Line Estimation.}
Prior work~\cite{li2023understanding} shows that the referent typically lies along the line connecting a person’s head and pointing finger. 
Following this intuition, we use the OpenPose~\cite{cao2017realtime} pose estimation model to detect approximate eye and fingertip coordinates, and draw a conic line from the detected eye to the fingertip.
This pointing line is then used in DADM to calculate distance for the final prediction. 

\subsection{Depth-Aware Decision Module}
We finally introduce a novel Depth-Aware Decision Module (DADM), see Algorithm \ref{alg:DADM}, which incorporates the top two predictions from $M_{aug}$ and $M_{depth}$ to determine the final output. 
The module first computes the intersection-over-union (IoU) between the highest-confidence bbox predictions of $M_{aug}$ and $M_{depth}$. 
If their overlap exceeds a threshold $T_1$, the final prediction is set to $b^0_{aug}$, since $M_{aug}$ generally demonstrates higher accuracy. 
If the overlap is below $T_1$, we instead use the predicted pointing line map and select the bounding box with the shortest distance to the line. 
Because the top predictions from both models may fail to overlap, especially in complex or ambiguous scenes, we also consider the second-ranked predictions. 
To avoid low-quality candidates, we include these only if their confidence score exceeds a second threshold $T_2$. 
A candidate list is then formed from the valid predictions, and the bbox whose center lies closest to the pointing line is chosen as the final output. 

\begin{algorithm}[t]
\caption{Depth-Aware Decision Module (DADM)}
\label{alg:DADM}
\begin{algorithmic}[1]
\Require Bounding box of Top-2 predictions from $M_{aug}$ and $M_{depth}$: $B_{aug} = [(b^0_{aug},c^0_{aug}), (b^1_{aug}, c^1_{aug})]$, $B_{depth} = [(b^0_{depth},c^0_{depth}), (b^1_{depth},c^1_{depth})]$
\Require Thresholds: $T_1 = 0.9$(IoU), $T_2 = 0.6$ (confidence)
\Require Predicted pointing line map $I_L$
\Ensure Final prediction $b^*$
 
\State Compute IoU: $iou \gets \text{IoU}(b^0_{aug}, b^0_{depth})$
\If{$iou \ge T_1$}
    \State $b^* \gets b^0_{aug}$
\Else
    \State Initialize candidate list: $\text{Candidates} \gets \{b^0_{aug}, b^0_{depth}\}$
    \For{each $(b^1, c^1) \in \{(b^1_{aug}, c^1_{aug}), (b^1_{depth}, c^1_{depth})\}$}
        \If{$c^1 \ge T_2$}
            \State Add $b^1$ to Candidates
        \EndIf
    \EndFor
    \State Compute distance of each candidate to $I_L$
    \State $b^* \gets$ candidate with shortest distance to $I_L$
\EndIf
\State \Return $b^*$
\end{algorithmic}
\end{algorithm}

\begin{figure*}[h!]
    \centering
    \includegraphics[width=0.79\linewidth]{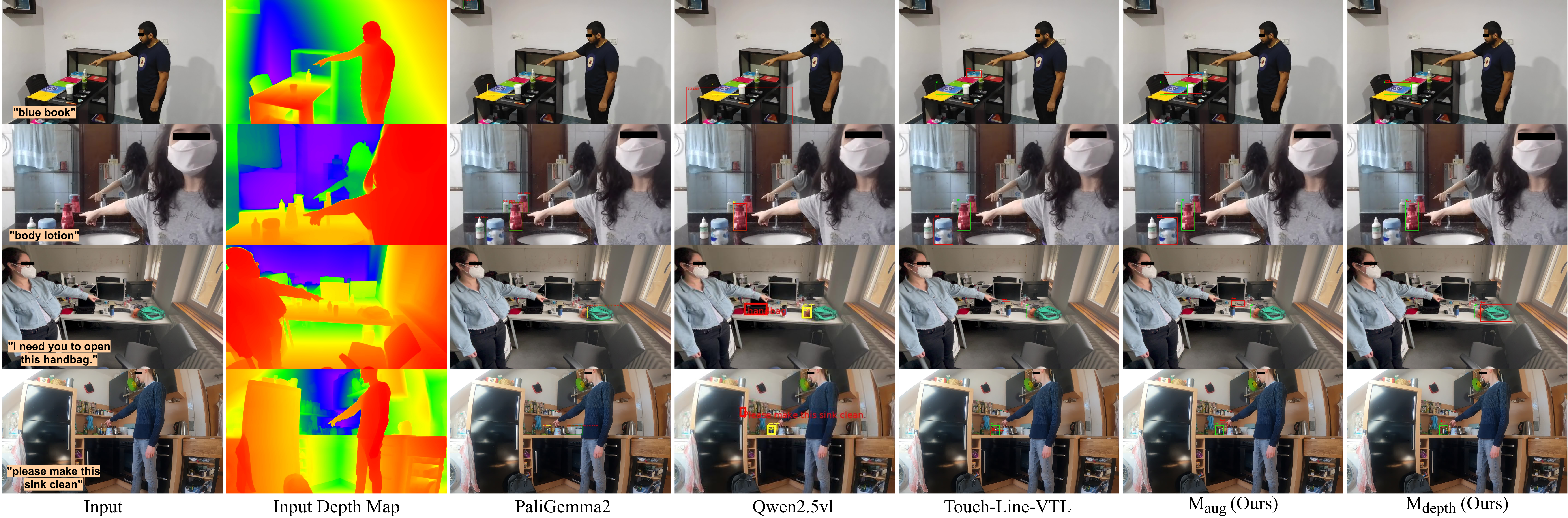}
    \caption{Visual comparisons on YouRefIt dataset (first two rows) and ISL pointing dataset (last two rows).}
    \label{fig:results}
\end{figure*}

\section{Experimental Results}
\label{sec:experiments}

\noindent\textbf{Datasets.}
For experiments, we utilize the YouRefIt~\cite{chen2021yourefit}, which contains 2970 training and 1251 test images. 
Further we evaluate our models on unseen ISL pointing dataset~\cite{constantin2022interactive} which contains ~190 video recordings.

\noindent\textbf{Evaluation.}
For evaluation metrics and setup, we follow prior work~\cite{chen2021yourefit}. 
IoU is computed at three threshold values: $0.25$, $0.50$, and $0.75$. 
Additionally, objects are categorized as \textit{small (S)}, \textit{medium (M)}, and \textit{large (L)} based on their bounding box size, and scores are reported with respect to object size.

\begin{table*}[t]
    \centering
    \footnotesize
    \begin{tabular}{l|cccc|cccc|cccc}
        \toprule
        IoU Threshold for mAP & \multicolumn{4}{c}{0.25} & \multicolumn{4}{c}{0.50} & \multicolumn{4}{c}{0.75} \\
        \midrule
        Object Sizes & \multicolumn{1}{c}{All} & \multicolumn{1}{c}{S} & \multicolumn{1}{c}{M} & \multicolumn{1}{c}{L} & \multicolumn{1}{c}{All} & \multicolumn{1}{c}{S} & \multicolumn{1}{c}{M} & \multicolumn{1}{c}{L} & \multicolumn{1}{c}{All} & \multicolumn{1}{c}{S} & \multicolumn{1}{c}{M} & \multicolumn{1}{c}{L} \\
        \midrule
        PaliGemma2 \cite{steiner2024paligemma}   & 58.8 & 29.0 & 53.5 & 75.8 & 46.9 & 22.1 & 50.8 & 68.0 & 31.7 & 6.2 & 34.1 & 54.8 \\
        Qwen2.5vl \cite{bai2025qwen2} & 38.9 & 17.0 & 41.8 & 58.0 & 31.0 & 11.1 & 33.6 & 48.1 & 20.0 & 5.7 & 19.8 & 34.5 \\
        Grounding DINO~\cite{liu2024grounding} & 57.9 & 38.0 & 60.9 & 74.9 & 54.9 & 35.7 & 59.3 & 69.6 & \textbf{42.3} & \textbf{22.7} & \textbf{45.9} & \underline{58.4} \\
        \midrule
        FAOA \cite{yang2019fast} & 44.5 & 30.6 & 48.6 & 54.1 & 30.4 & 15.8 & 36.5 & 39.3 & 8.5 & 1.4 & 9.6 & 14.4 \\
        ReSC \cite{yang2020improving}  & 49.2 & 32.3 & 54.7 & 60.1 & 34.9 & 14.1 & 42.5 & 47.7 & 10.5 & 0.2 & 10.6 & 20.1 \\
        YourRefit PAF \cite{chen2021yourefit} & 52.6 & 35.9 & 60.5 & 61.4 & 37.6 & 14.6 & 49.1 & 49.1 & 12.7 & 1.0 & 16.5 & 20.5 \\
        YourRefit Full \cite{chen2021yourefit} & 54.7 & 38.5 & 64.1 & 61.6 & 40.5 & 16.3 & 54.4 & 51.1 & 14.0 & 1.2 & 17.2 & 23.3 \\
        REP \cite{shi2022spatial} & 58.8 & 44.7 & 68.9 & 63.2 & 45.7 & 25.4 & 57.7 & 54.3 & 18.8 & 3.8 & 22.2 & 29.9 \\
        Touch-Line-EWL \cite{li2023understanding} & {69.5} & {56.6} & {71.7} & 80.0 & 60.7 & {44.4} & 66.2 & 71.2 & 35.5 & {11.8} & {38.9} & 55.0 \\
        Touch-Line-VTL \cite{li2023understanding} & 71.1 & 55.9 & 75.5 & {81.7} & {63.5} & 47.0 & \underline{70.2} & 73.1 & \underline{39.0} & \underline{13.4} & \underline{45.2} & {57.8} \\
        \midrule
        Baseline  & 71.2 & 59.5 & 73.0 & 80.8 &	60.1 & 43.2 & 66.4 & 70.5 &	32.8 & 8.6 & 35.4 & 53.4 \\ 
        $M_{aug}$ (Only Aug)   & \underline{73.2} & \underline{60.7} & \underline{75.7} & \underline{83.5} & \underline{64.1} & \underline{47.5} & 69.8 & \underline{75.3} & 35.2 & 12.7 & 37.2 & 55.8 \\
        $M_{depth}$ (Only Depth) & 70.8	& 56.5 & 73.3 & 82.7 & 60.8 & 43.5 & 67.9 & 71.4 & 33.0 & 9.5 & 37.7 & 52.1  \\
        $M_{aug\_depth}$ (Aug + Depth) & 66.1	& 56.1 & 69.7 & 72.4 & 53.6 & 37.9 & 61.7 & 61.2 & 23.3 & 6.5 & 22.0 & 40.7  \\
        DA-ERU (Full)       & \textbf{78.7} & \textbf{65.7} & \textbf{82.1} & \textbf{88.4} & \textbf{67.6} & \textbf{48.5} & \textbf{75.4} & \textbf{79.3} & 38.1 & \underline{13.4} & 39.8 & \textbf{61.0}  \\
        \bottomrule
    \end{tabular}
    \caption{Comparison of our model with prior work in terms of mean Average Precision (mAP) at different IoU thresholds, across various object sizes, on the YouRefIt dataset~\cite{chen2021yourefit}.}
    \label{tab:comparison_iou}
\end{table*}

\noindent\textbf{Results.}
We present quantitative results in Table \ref{tab:comparison_iou}.
The first section of the table reports results from recent LMMs~\cite{steiner2024paligemma, bai2025qwen2} and the SOTA open-set object detector Grounding DINO~\cite{liu2024grounding}. 
We evaluate these models without finetuning to leverage their zero-shot capabilities.
The second section summarizes results from recent prior works on this task. 
The third section presents the performance of our models. 
Here, \textit{baseline} refers to straightforward training of our architecture with YouRefIt dataset;
  \textit{Only Aug} corresponds to $M_{aug}$ trained with text augmented data; \textit{Only Depth} shows the performance of $M_{depth}$; and \textit{Full} represents the combination of $M_{aug}$ and $M_{depth}$ with the DADM.
Our baseline model achieves competitive performance compared to the previous model~\cite{li2023understanding}. 
The $M_{aug}$ model clearly outperforms it at the $0.25$ and $0.50$ IoU thresholds. 
Although $M_{depth}$ is less accurate than the augmented model overall, analysis of its outputs reveals that in critical cases, where most models fail due to missing depth information, it performs robustly thanks to depth awareness. 
This motivates training $M_{depth}$ with augmented data, however, it harms the performance.  
Therefore, we introduce DADM, we obtain a substantial performance boost and achieve SOTA performance with \textit{Full} model, except for \textit{small} and \textit{medium} objects at $0.75$ threshold.
The evaluated LMMs perform significantly worse than our proposed method. 
We further evaluate our full model on the unseen and more challenging ISL pointing dataset \cite{constantin2022interactive}. 
The results show that our model surpasses both Touch-Line models and LMMs by a large margin at $IoU=0.25$ and $IoU=0.50$ thresholds, while achieving comparable performance at $IoU = 0.75$. 
Note that the $0.75$ threshold primarily reflects the precision of predicted bounding box coordinates. 
Our model is more accurate in identifying the correct target object in the scene, though it occasionally produces slightly less precise bounding boxes. 
In Fig. \ref{fig:results}, we present qualitative results together with the predicted depth maps.

\begin{table}[]
    \centering
    \footnotesize
    \begin{tabular}{c|ccc}
        \toprule
        Setup & IoU=0.25 & IoU=0.50 & IoU=0.75 \\
        \midrule
        PaliGemma2 \cite{steiner2024paligemma} &     47.2 & 39.5 & \textbf{31.6} \\
        Qwen2.5vl \cite{bai2025qwen2} & 32.5 & 32.1 & 29.2 \\
        Touch-Line-EWL \cite{li2023understanding} & 45.0 & 35.8 & 22.0 \\ 
        Touch-Line-VTL \cite{li2023understanding} & 47.7 & 36.7 & 17.4  \\
        \midrule
        DA-ERU (Full)  & \textbf{62.7} & \textbf{50.1} & 30.1 \\
        \bottomrule
    \end{tabular}
    \caption{Test results on unseen ISL pointing dataset~\cite{constantin2022interactive}.}
    \label{tab:comparison_iou_isl}
\end{table}

\noindent\textbf{Ablation Study for Decision Method in DADM.}
The first part of Table \ref{tab:ablation_DADM_method} presents an ablation study of different decision strategies within DADM. 
\texttt{A} selects the top-1 predictions from both models and chooses the box with the shortest distance to the pointing line. 
\texttt{B}, instead, selects the box with the largest overlap in pixels with the pointing line. 
Finally, \texttt{C} normalizes this overlap by bounding-box size and selects the box with the highest percentage. 
While all methods yield performance gains, our proposed DADM achieves the most significant improvement.

\begin{table}[]
    \centering
    \footnotesize
    \resizebox{\linewidth}{!}{
    \begin{tabular}{l|ccc}
        \toprule
        Setup & IoU=0.25 & IoU=0.50 & IoU=0.75 \\
        \midrule 
        \texttt{A} - Distance to Pointing Line & 75.5 & 64.8 & 35.8 \\
        \texttt{B} - Overlapping area w/ Pointing Line & 75.9 & 64.7 & 35.9 \\
        \texttt{C} - Overlapping area w/ Pointing Line Percentage & 75.1 & 64.4 & 35.9 \\
        \midrule
        Confidence score based  & 73.3 & 63.8 & 36.3 \\
        Adaptive depth & 73.3 & 63.0 & 35.9 \\
        \midrule
        DADM (Distance) & \textbf{78.7} & \textbf{67.6} & \textbf{38.1} \\
        \bottomrule
    \end{tabular}}
    \caption{Ablation study for DADM and distance strategy.}
    \label{tab:ablation_DADM_method}
\end{table}

\noindent\textbf{Ablation Study for Final Decision.}
The second part of Table \ref{tab:ablation_DADM_method} compares alternative strategies to DADM for selecting the final object. 
The first confidence score based method selects between the top-1 predictions of both models by choosing the box with the higher confidence score. 
Adaptive depth also uses the top-1 predictions, but selects the model whose confidence gap between the first and second prediction is larger, indicating a clearer decision. 
Among these strategies, DADM consistently achieves the best performance.

\section{Conclusion}
We address limitations of existing ERU methods with the following three contributions: text augmentation, depth estimation, and the DADM module. 
Experiments show that text augmentation and depth maps each improve performance.
Their combination with our DADM, which leverages predicted pointing lines for distance-based selection, yields more accurate and robust understanding in complex visual scenes.
We validated our contributions on two benchmark datasets.

\noindent\textbf{Acknowledgment.}
This work was supported in part by the EU’s Horizon research and innovation program, projects: Meetween (101135798), DVPS (101213369). Development of tools and applications developed under contract from KIT Campus Transfer GmbH (KCT).




\bibliographystyle{IEEEbib}
\bibliography{strings,refs}

\end{document}